\documentclass{article}

\usepackage[dblblindworkshop, final]{neurips_2026}
\workshoptitle{Continual Learning for Enterprise AI Agents (CLEA)}

\usepackage[T1]{fontenc}
\usepackage{amsfonts}
\usepackage{amsmath}
\usepackage{amssymb}
\usepackage{booktabs}
\usepackage{caption}
\usepackage{float}
\usepackage{graphicx}
\usepackage{listings}
\usepackage{lineno} 
\usepackage{microtype}
\usepackage{tabularx}
\usepackage{url}
\usepackage{xcolor}

\definecolor{linkcol}{HTML}{1F6E9C}
\usepackage[colorlinks=true,linkcolor=linkcol,citecolor=linkcol,
            urlcolor=linkcol,breaklinks=true]{hyperref}

\newcommand{\fico}{\textsc{FiCo}}
\newcommand{\vigil}{\textsc{Vigil}}
\newcolumntype{Y}{>{\raggedright\arraybackslash}X}
\newcolumntype{C}{>{\centering\arraybackslash}X}

\providecommand{\FicoDBFullFinal}{21.3\%}

\providecommand{\FicoDBNoSummaryFinal}{18.7\%}

\providecommand{\FicoDBNoRerankFinal}{20.4\%}

\providecommand{\FicoDBNoFallbackFinal}{21.3\%}

\providecommand{\FicoDBGoldDenotation}{25.9\%}
\providecommand{\FicoDBGoldFinal}{27.8\%}

\providecommand{\FicoDBGoldNoSamplesDenotation}{18.3\%}
\providecommand{\FicoDBGoldNoSamplesFinal}{20.8\%}

\providecommand{\CLStaticForward}{75.8}
\providecommand{\CLUngatedForward}{82.4}
\providecommand{\CLGuardedForward}{82.0}
\providecommand{\CLStaticCoverage}{87.1}
\providecommand{\CLUngatedCoverage}{94.2}
\providecommand{\CLGuardedCoverage}{93.8}
\providecommand{\CLStaticPrecision}{87.0}
\providecommand{\CLUngatedPrecision}{87.5}
\providecommand{\CLGuardedPrecision}{87.5}
\providecommand{\CLUngatedDelta}{6.7}
\providecommand{\CLGuardedDelta}{6.3}
\providecommand{\CLGuardedDeltaLow}{5.6}
\providecommand{\CLGuardedDeltaHigh}{6.9}
\providecommand{\CLUngatedBWT}{2.3}
\providecommand{\CLGuardedBWT}{2.7}

\providecommand{\CLHarmfulProposals}{8}
\providecommand{\CLCheckedProposals}{40}

\providecommand{\CLCheckedPromotions}{16}
\providecommand{\CLActionOracle}{84.4}

\providecommand{\CLAuditQuestions}{380}
\providecommand{\CLAuditDocuments}{16}
\providecommand{\CLAuditStaticAccuracy}{80.3}
\providecommand{\CLAuditCalibratorAccuracy}{86.2}
\providecommand{\CLAuditSelectorAccuracy}{80.5}
\providecommand{\CLAuditUngatedAccuracy}{86.6}
\providecommand{\CLAuditAccuracyGateAccuracy}{86.7}
\providecommand{\CLAuditDualGateAccuracy}{86.5}
\providecommand{\CLAuditAccuracyGateDelta}{6.4}
\providecommand{\CLAuditAccuracyGateCoverage}{96.8}
\providecommand{\CLAuditAccuracyGatePrecision}{89.6}
\providecommand{\CLAuditPromoted}{26}
\providecommand{\CLAuditPromotedHarmful}{3}
\providecommand{\CLAuditRejected}{24}
\providecommand{\CLAuditRejectedHarmful}{6}

\newcommand{\OracleTypeOOFAccuracy}{89.9}
\newcommand{\OracleTypeAuditAccuracy}{91.1}
\newcommand{\OracleGoldDocGoldTypeAccuracy}{76.3}
\newcommand{\OracleSelectedDocGoldTypeAccuracy}{66.3}
\newcommand{\OracleGoldDocPredictedTypeAccuracy}{70.1}
\newcommand{\OracleSelectedDocPredictedTypeAccuracy}{60.8}
\newcommand{\OracleTypeCostGoldDoc}{6.2}
\newcommand{\OracleTypeCostGoldDocLow}{4.6}
\newcommand{\OracleTypeCostGoldDocHigh}{7.9}

\newcommand{\OracleBridgeGap}{15.5}
\newcommand{\OracleBridgeGapLow}{12.7}
\newcommand{\OracleBridgeGapHigh}{18.5}
\newcommand{\OracleZeroShotTypeAccuracy}{72.3}
\newcommand{\OracleZeroShotBridgeAccuracy}{49.7}
\newcommand{\OracleLearnedOverZeroShotBridge}{11.2}

\providecommand{\CLBSpectrumVigil}{55.605}
\providecommand{\CLBSpectrumPrior}{46.198}
\providecommand{\CLBSpectrumDelta}{9.407}
\providecommand{\CLBSalesVigil}{11.202}
\providecommand{\CLBSalesPrior}{10.073}
\providecommand{\CLBSalesDelta}{1.129}
\providecommand{\CLBPokerVigil}{346.400}
\providecommand{\CLBPokerPrior}{343.020}
\providecommand{\CLBPokerDelta}{3.380}

\newcommand{\MiMoDirectMean}{85.66}

\newcommand{\MiMoSemanticMean}{85.70}

\newcommand{\MiMoSemanticChangedDirect}{61.43}
\newcommand{\MiMoSemanticChangedAccuracy}{66.43}

\newcommand{\CLUngatedRetrospective}{\CLUngatedBWT}
\newcommand{\CLGuardedRetrospective}{\CLGuardedBWT}

\begin{document}

\title{\vigil{}: Verifier-Informed Gated Improvement Loop for Continual Learning in Enterprise Agents}
\author{%
  Kang Li \\
  Atlassian \\
  \texttt{kli3@atlassian.com} \\
  \And
  Lu He \\
  Atlassian \\
  \texttt{lhe@atlassian.com} \\
  \And
  Sandarsita Guntupalli \\
  Carnegie Mellon University \\
  \texttt{sandarsg@andrew.cmu.edu}
}
\maketitle

\begin{abstract}
Enterprise agents should improve from delayed feedback without allowing every correction to rewrite system behavior. We study continual harness learning for corpus-level spreadsheet question answering. Building on \fico{} (Find-then-Compute), a static retrieval-and-execution backbone, we introduce \vigil{} (Verifier-Informed Gated Improvement Loop). Within a question, \vigil{} verifies and repairs diversely prompted Structured Query Language (SQL) candidates. Across episodes, delayed labels update only a contract calibrator and query/result-column selector. The base model, prompts, retriever, and recorded candidate pool remain fixed in the continual-learning protocol. With the gold workbook and expected type supplied, the ungated and dual-gated full-replay variants reach 82.4\% and 82.0\% forward accuracy over 79 documents, from a 75.8\% static baseline. The dual gate has the larger retrospective gain (2.7 versus 2.3 points), and its 6.3-point forward gain has a 95\% $t$-interval of 5.6--6.9. In a separate stricter split that excludes 16 documents and 380 questions from fitting and online promotion, the accuracy-only gate raises mean held-out accuracy across ten final harnesses from \CLAuditStaticAccuracy\% to \CLAuditAccuracyGateAccuracy\%. Calibrator-only adaptation gains 5.9 points, close to the combined 6.4-point gain. Yet three of 26 gate-approved updates reduce held-out accuracy relative to their incumbents, so replay-buffer non-regression does not imply held-out non-regression. MiMoTable and external-task case studies test within-task verification and reuse the same promote-or-retain discipline. Overall, the results support bounded, auditable harness adaptation while revealing where finite replay gates fail to generalize beyond their promotion buffers.\looseness=-1
\end{abstract}

\section{Introduction}

Enterprise agents are deployed as systems rather than isolated models. Their behavior depends on retrieval, prompts, tools, verifiers, routing, and answer policies. As schemas and workflows evolve, delayed corrections should improve this harness without eroding behavior that users already rely on. Updating foundation-model weights may be unavailable or unnecessarily broad. Unconstrained self-modification is also difficult to audit. We therefore study a bounded form of continual learning in which labeled trajectories update only explicit harness components and every proposed update must pass a document-disjoint replay gate.

Corpus-level spreadsheet question answering (QA) is a demanding enterprise testbed for this problem. The agent must identify a workbook and sheet, express an operation, execute it, and decide whether the result satisfies the request. Retrieval-augmented generation (RAG) works when answers occur in retrievable passages~\citep{lewis2020retrieval}, but spreadsheet questions often require counts, comparisons, or grouped aggregation over many rows. A bounded text context may expose the schema while omitting cells needed for the answer. Execution can still succeed with an incorrect filter, grouping, or aggregate.\looseness=-1

Our operational reference point is a prefix-RAG configuration that serializes inferred headers and at most three data rows per sheet, truncates the representation to 2{,}000 tokens, retrieves by embedding similarity, and asks the language model to answer from the retrieved text. It is an engineering reproduction of a deployed configuration rather than a competitive gold-table oracle. It exposes a common enterprise failure mode: the correct workbook can be retrieved even though the rows needed for computation were never indexed.

We build on \fico{} (Find-then-Compute), a static spreadsheet agent that retrieves workbook summaries, reranks candidates using schema information, and executes a constrained query over the selected full table. \vigil{} adds answer-stage verification, structured repair, persistent trajectory logging, and replay-gated harness updates around this frozen backbone.

We make four contributions. First, we formulate enterprise-agent improvement as continual learning over an explicit harness. Delayed labels update contract calibration and action selection while the base model, prompts, retriever, and non-relaxable execution checks remain frozen. Second, \vigil{} introduces a transactional update protocol in which immutable trajectories make proposals replayable and a document-disjoint gate makes promotion rejectable and auditable. Third, two complementary replay analyses separate forward transfer from generalization to a fixed held-out document set and expose residual regression risk. Fourth, controlled spreadsheet experiments quantify repair and oracle costs. MiMoTable and external-task case studies test the same promote-or-retain design in other settings.

\section{Related Work}

\paragraph{Spreadsheet agents and program verification.}
Table question answering and natural-language database-querying benchmarks usually provide the relevant table or database~\citep{pasupat2015compositional,yu2018spider,li2023bird,oses2024databench}. Prior systems compress workbook structure~\citep{tian2024spreadsheetllm}, retrieve from table corpora~\citep{pan2022trag,zou2025trag}, and search multimodal workbooks~\citep{gulati2026frtr,gulati2026brtr}. Most closely related, TableRAG couples retrieval with structured query execution over heterogeneous text and tables~\citep{yu2025tablerag}. Constrained decoding prevents invalid programs~\citep{scholak2021picard}, while execution-based reranking verifies sampled programs~\citep{ni2023lever}. \fico{} addresses blind workbook selection and whole-table execution, while \vigil{} adds contract-based answer verification and replay-gated adaptation around a frozen, auditable backbone.

\paragraph{Continual learning for agent harnesses.}
Lifelong-agent work studies adaptation through memory, action policies, and feedback~\citep{zheng2025lifelong}. Reflexion stores linguistic feedback in memory~\citep{shinn2023reflexion}. Reflective prompt evolution uses trajectories to optimize prompts~\citep{agrawal2025gepa}. Continual skill learning can improve reusable workflows but may drift under self-feedback~\citep{zhong2026skilllearnbench}. \vigil{} does not claim replay, verification, or frozen-model adaptation as individually new. Its design contribution is a transactional harness protocol: only the calibrator and selector are mutable, trajectories are immutable, fitting and promotion documents are disjoint, and a fixed held-out audit measures transfer beyond the promotion buffer. Unlike memory or prompt evolution, generation stays frozen and every update has an explicit promote-or-retain decision. Enterprise evaluation must itself evolve~\citep{saxena2025continuous}.

\section{Task and Method}
\label{sec:method}

Let a collection contain documents $\mathcal{D}=\{d_1,\ldots,d_n\}$, each with one or more tables. Given a question $q$, a system selects a document and table, generates an executable operation, executes it, and returns or abstains from an answer. End-to-end accuracy therefore depends jointly on retrieval, target selection, program semantics, verification, and normalization.

\subsection{\fico{}: Static Find-then-Compute Backbone}
\label{sec:fico-backbone}

\paragraph{Find: summary retrieval and schema reranking.}
We support Microsoft Excel workbooks, Google Sheets, comma-separated values (CSV), JavaScript Object Notation (JSON), and Parquet. At ingestion time, each source is normalized into a shared SQLite database. Single-table formats contribute one table, while multi-tab spreadsheets contribute one table per tab. For irregular spreadsheets, we infer the true header row by scoring candidate rows according to column count and cell density. We also reshape wide tables that encode years as columns into a standard long format for querying. A language model then produces a concise summary of each source that describes its topic, tab contents, and column names. We embed one summary per source using \texttt{embeddinggemma-300m}, retrieve the ten most relevant candidates, and rerank them using a \emph{schema signature} composed of column names and representative values. If the highest-ranked candidate returns an empty result, the system may try the next candidate. This fallback can recover from an incorrect initial selection, but it can also direct execution to the wrong source.

\paragraph{Compute: constrained whole-table execution.}
For the selected table, the prompt lists row count, numeric ranges or representative values, and up to three sample rows. Gemini~3~Flash generates a Structured Query Language (SQL) program that is executed locally over the full table. The deployed \fico{} end-to-end track uses SQLite, feeds execution errors back to the model for up to two repair attempts, and converts the result to the benchmark's expected answer form.

\subsection{\vigil{}: Verifier-Informed Gated Improvement Loop}
\label{sec:vigil-loop}

\paragraph{Harness state and transactional updates.}
\vigil{} wraps the static \fico{} backbone with learned verification and transactional adaptation. We represent the resulting harness as
\begin{equation}
\mathcal{H}_{\theta}=(R,G,E,C_{\mathrm{hard}},V_{\theta},S_{\theta}),
\label{eq:harness}
\end{equation}
where $R$ retrieves and loads tables, $G$ generates candidate programs, $E$ executes them, and $C_{\mathrm{hard}}$ enforces non-learned admissibility checks. $V_{\theta}$ calibrates answer-contract acceptance, and $S_{\theta}$ selects among accepted query/result-column actions. The vector $\theta$ denotes their jointly mutable parameters. In the continual protocol, $R$, $G$, $E$, $C_{\mathrm{hard}}$, and the generated candidate pool are fixed. Once a delayed label $y_i$ arrives, question $i$ yields an immutable trajectory
$\tau_i=(d_i,q_i,\mathcal{P}_i,Z_i,I_i,a_i,y_i)$ containing its candidate programs, denotations, verifier issues, selected action, and label. Replaying the same $\tau_i$ under two harness states therefore changes only calibration or selection, not the evidence being scored.

The contribution is not a new SQL decoder, replay buffer, or verifier in isolation. \vigil{} couples four restrictions into one learning interface: a frozen proposal distribution, explicitly bounded mutable state, immutable decision evidence with document-disjoint fitting and promotion sets, and an atomic promote-or-retain transition. This construction turns delayed feedback into a rejectable harness transaction. Unlike approaches that rewrite prompts, memories, or model weights, a proposal can be evaluated against the incumbent on exactly the same actions and denotations. A rejected proposal changes no deployed state. The design consequently supports paired attribution of gains and regressions to $\theta$, while preserving a narrow set of non-learned execution invariants.

\paragraph{Verification, selection, and repair.}
Execution success does not establish semantic correctness. A query can run while using the wrong filter, aggregation, grouping, or result column. In the controlled \vigil{} track, the system generates two diversely prompted SQL candidates and executes both in DuckDB. A candidate is accepted only if it is a single read-only statement, executes successfully, satisfies the supplied answer-type cardinality, exposes exactly one compatible result column, and passes intent checks for counts, averages, medians, extrema, and explicit top-$k$ requests. Candidates are ranked by contract satisfaction, with shorter SQL used as a tie-breaker. If neither initial candidate is accepted, the most informative failure and its exact parser, binder, execution, or contract feedback are supplied for up to two repair rounds. If no repair passes, the system abstains.

An answer contract specifies the expected semantic type and cardinality of an output. The verifier maps a selected result to Boolean, number, category, list of numbers, or list of categories. Scalars require one row and one compatible answer column, whereas lists preserve multiplicity. These constraints define structural admissibility before selection. Satisfying them does not by itself establish that the query used the correct filter, aggregation, or scope.

\paragraph{Two adaptation timescales.}
\vigil{} separates within-question repair from across-episode learning. Within a question, $G$ produces two diversely prompted candidates. The system changes only the candidate program through at most two repair rounds, using structured failure evidence from $E$, $C_{\mathrm{hard}}$, and $V_{\theta}$. It does not update persistent state. Across episodes, delayed labels fit a regularized calibrator and selector using answer type, candidate stage, SQL operators, verifier issue class, result-column position, and lexical column cues. Let $F_t$ be fitting documents and $B_t$ the promotion buffer at episode $t$, with $F_t\cap B_t=\emptyset$. A proposal is
$\theta'_t=\operatorname{Fit}(\theta_t,\{\tau_i:d_i\in F_t\})$ and is evaluated on the same buffered trajectories as the incumbent. The dual gate used in full replay is
\begin{equation}
g_t=\mathrm{I}\!\left[
\operatorname{Acc}(\theta'_t;B_t)>\operatorname{Acc}(\theta_t;B_t)
\ \land\
\operatorname{SelAcc}(\theta'_t;B_t)\geq\operatorname{SelAcc}(\theta_t;B_t)
\right],
\label{eq:gate}
\end{equation}
where $\mathrm{I}$ is the indicator function, $\operatorname{Acc}$ counts abstentions as incorrect, and $\operatorname{SelAcc}$ is accuracy among accepted outputs. The deployed state then changes atomically:
\begin{equation}
\theta_{t+1}=g_t\theta'_t+(1-g_t)\theta_t.
\label{eq:transaction}
\end{equation}
The held-out analysis additionally evaluates an accuracy-only gate by removing the second condition. Persistent state consists only of $\theta_t$, promotion history, and immutable labeled trajectories. The retriever, loader, language model, generation prompts, and candidate pool remain frozen.

For every $\theta$, $C_{\mathrm{hard}}$ rejects candidates with parser, binder, or execution failures, as well as candidates that violate the result cap, single-statement, mutation, or read-only constraints. Because $C_{\mathrm{hard}}$ is outside $\theta$, promotion cannot relax these checks. Because Equation~\ref{eq:transaction} is atomic, rejection exactly retains the incumbent parameters. These are deterministic properties of the reported state machine, not claims of complete production security. SQL validation does not use a complete parser, execution lacks an explicit timeout, and cell-level prompt injection is not addressed. Equation~\ref{eq:gate} establishes only paired non-regression on the finite buffer $B_t$. It does not guarantee semantic correctness, distribution-shift robustness, or performance on future documents.

Table~\ref{tab:mutable} makes the adaptation boundary explicit. Within-question repair changes only candidate SQL. Across episodes, learning changes only contract calibration and query/result-column selection. The base model, prompts, retriever, loader, candidate pool, and hard execution checks stay fixed.

\begin{table}[t]
\centering
\footnotesize
\setlength{\tabcolsep}{4pt}
\begin{tabularx}{\textwidth}{Y C C}
\toprule
\textbf{Component} & \textbf{Within-question repair} & \textbf{Across-episode learning} \\
\midrule
Workbook retriever and loader & Frozen or oracle-supplied & Frozen or oracle-supplied \\
Base model and generation prompts & Frozen & Frozen \\
Candidate SQL & Repairable & Frozen replay pool \\
Answer-contract calibrator & Frozen & Learnable \\
Query/result-column selector & Frozen & Learnable \\
Parser, execution, and hard checks & Non-relaxable & Non-relaxable \\
\bottomrule
\end{tabularx}
\caption{Mutable scope of the reported adaptation experiments. Learning does not regenerate candidates or update the base model, prompts, retriever, or loader.}
\label{tab:mutable}
\end{table}

\begin{figure}[t]
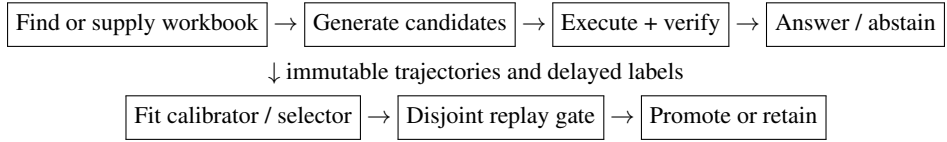

\centering
\small
\fbox{\strut Find or supply workbook} $\rightarrow$
\fbox{\strut Generate candidates} $\rightarrow$
\fbox{\strut Execute + verify} $\rightarrow$
\fbox{\strut Answer / abstain}
\\[5pt]
$\downarrow$ immutable trajectories and delayed labels \\[3pt]
\fbox{\strut Fit calibrator / selector} $\rightarrow$
\fbox{\strut Disjoint replay gate} $\rightarrow$
\fbox{\strut Promote or retain}
\caption{\vigil{} at two timescales. The upper path performs bounded within-question repair. The lower path updates contract calibration and query/result-column selection from delayed labels, then applies a replay gate to a frozen action pool. The end-to-end track includes workbook retrieval. The verifier and continual tracks begin with a gold workbook and expected-type metadata.}
\label{fig:loop}
\end{figure}

\section{Experiments}
\label{sec:experiments}

\subsection{Research questions and evaluation ledger}

We study four core research questions. \textbf{Research Question 1 (RQ1)} tests \fico{} end-to-end value. \textbf{Research Question 2 (RQ2)} localizes errors and component effects. \textbf{Research Question 3 (RQ3)} measures \vigil{} verification, repair, and removal of controlled inputs, including the official-workbook MiMoTable case study. \textbf{Research Question 4 (RQ4)} tests continual transfer and regression control in both full replay and a stricter fixed held-out audit. Appendix~\ref{app:clbench} reports separate cross-domain promote-or-retain case studies. Table~\ref{tab:ledger} fixes each comparison's scope. Absolute scores are compared only within a row.

\subsection{Benchmarks, evaluation tracks, and metrics}

The DataBench bundle contains 80 datasets, 3{,}708{,}884 rows, and 1{,}810 questions spanning Boolean, numeric, categorical, and list-valued answers. It extends the original 65-dataset benchmark~\citep{oses2024databench,oses2025semeval}. The corpus-retrieval track also evaluates 508 English questions over 174 multi-sheet MiMoTable workbooks~\citep{li2025mimotable}. Prefix-RAG and \fico{} share the embedding model, retrieval depth, Gemini~3~Flash generator, and production-style judge. The controlled verifier and continual tracks instead use DuckDB, the gold workbook, and DataBench's five-way answer-type field. Because participants do not receive this field~\citep{oses2025semeval}, we treat it as an oracle and do not claim deployment-level verifier accuracy. Deterministic equality preserves list multiplicity, applies the supplied 1\% numeric tolerance, normalizes Booleans exactly, and canonicalizes whitespace, case, and equivalent date serializations. The workbook/type bridge varies both controlled inputs, and absolute scores are not compared across runners.

For the official-workbook MiMoTable case study, we pin the authors' repository at commit \texttt{96ff0bc} and preserve its 641 answer-bearing English text-only questions and 174 complete workbooks. A static adapter renders every nonempty cell in the benchmark's markdown convention and answers with GPT-5.4 at medium reasoning. A gold-blind, high-reasoning verifier independently recomputes the result and proposes \textsc{retain} or \textsc{replace}. For a replacement, another high-reasoning call answers the unchanged official prompt. A medium-reasoning gate promotes the replacement only when the verifier's proposed answer and the independent solve are semantically equivalent. Ambiguity or request failure retains the incumbent. References and judge verdicts are absent from every decision. Following the released protocol, GPT-4o (\texttt{gpt-4o-2024-08-06}, temperature zero) judges predictions against references. We repeat the complete 641-question judging run three times and report its range and mean. The 79.0\% Claude~3.5~Sonnet comparator is reported rather than rerun. This case study tests within-task verification, not continual learning.

\begin{table}[t]
\centering
\scriptsize
\setlength{\tabcolsep}{2.5pt}
\begin{tabularx}{\textwidth}{l l Y Y r Y}
\toprule
 & \textbf{Track} & \textbf{Systems / intervention} & \textbf{Oracle / controlled input} & \textbf{$n$} & \textbf{Valid claim} \\
\midrule
RQ1 & End to end & Prefix-RAG compared with \fico{} & None & 1{,}810 / 508 & Shared-judge end-to-end difference \\
RQ2 & Controlled ablation & Single-component removal & Gold workbook (compute rows) & 1{,}810 & Paired effect in fixed runner \\
RQ3 & Verifier / bridge & Repair ladder and workbook/type $2\times2$ & Varied by arm & 1{,}810 & Repair and oracle effects \\
RQ3 & Official MiMoTable & Direct compared with semantic gate & Benchmark-supplied workbook & 641 & Official-prompt judged accuracy \\
RQ4a & Full replay & Static, ungated, and dual-gated & Workbook + type, delayed labels & $1{,}810\times10$ & Forward and retrospective gain \\
RQ4b & Held-out audit & Static and five adaptive variants & Workbook + type, delayed labels & $1{,}430$ per ordering + 380 audit & Held-out transfer and gate behavior \\
\bottomrule
\end{tabularx}
\caption{Evaluation ledger. Changing runner, execution engine, oracle, or evaluator defines a new track. Gold-workbook treatments are not end-to-end competitors. RQ4a and RQ4b use different partitions. RQ4b learns from 1,430 questions per ordering and evaluates on 380 fixed audit questions.}
\label{tab:ledger}
\end{table}

Paired controlled interventions remove summaries, schema reranking, empty-result fallback, or---under a gold-workbook oracle---representative values and sample rows. Gold-workbook trials still select the sheet on multi-sheet workbooks. We report three answer metrics. \emph{Coverage} is the fraction of questions with an accepted output, \emph{selective accuracy} is correctness among accepted outputs, and overall accuracy counts abstentions as incorrect.

The continual-learning evaluation uses two distinct partitions. Full replay places all 79 traced document identifiers, covering 1{,}810 question trajectories, into five question-balanced episodes under ten orderings. The eightieth DataBench bundle entry has no question trajectories. Its dual gate requires positive promotion-buffer accuracy gain and no selective-accuracy loss. The stricter analysis fixes 16 documents and 380 questions before any ordering is generated. These documents never enter fitting, threshold selection, replay gating, or promotion. The remaining 63 documents and 1{,}430 questions form five learning episodes. Final harnesses are compared on the fixed held-out set under static, calibrator-only, selector-only, combined ungated, accuracy-only-gated, and dual-gated treatments.

Each learning episode is evaluated before its labels are exposed. Previously observed documents are split by stable hash into disjoint 70\% fitting and 30\% promotion sets. Let $A_{t,j}$ denote accuracy on episode $j$ after updates through episode $t$. Forward accuracy is $\frac{1}{T-1}\sum_{j=2}^{T} A_{j-1,j}$. Retrospective gain is $\frac{1}{T-1}\sum_{j=1}^{T-1}(A_{T,j}-A_{j-1,j})$, comparing final with pre-exposure performance on earlier episodes. It is not standard backward transfer. Intervals are 95\% $t$-intervals over paired per-ordering differences.

Finally, a 100-question diagnostic over 20 spreadsheets compares four textual representations at a fixed top-5 budget. Coverage@5 requires retrieved units to expose the values or sufficient statistics needed to answer. Seventy-three questions require computation, and the best observed textual coverage is 39\% (Appendix Table~\ref{tab:chunking}). This result is budget-specific, not a universal retrieval ceiling.

\section{Results and Analysis}
\label{sec:results}

\subsection{RQ1: End-to-end system value}

In the corpus-retrieval setting, \fico{} improves over Prefix-RAG by 63.4 points on DataBench and 57.5 points on the 508-question MiMoTable subset (Table~\ref{tab:endtoend}). On DataBench, \fico{} is correct on 1{,}172 questions for which the baseline is wrong, while the reverse holds on 24. A continuity-corrected McNemar test over 1{,}196 discordant pairs gives $\chi^2=1100.0$ ($p<10^{-6}$). This is an operational comparison, not a leaderboard claim against gold-workbook systems.

\begin{table}[H]
\centering
\small
\begin{tabular}{lrr}
\toprule
\textbf{System} & \textbf{DataBench} & \textbf{MiMoTable retrieval} \\
\midrule
Prefix-RAG baseline & 12.8\% & 22.2\% \\
\fico{} & \textbf{76.2\%} & \textbf{79.7\%} \\
\midrule
$\Delta$ & $+63.4$ & $+57.5$ \\
\bottomrule
\end{tabular}
\caption{End-to-end accuracy for corpus-selection tracks. $\Delta$ is measured in percentage points.}
\label{tab:endtoend}
\end{table}

\subsection{RQ2: Error localization and component effects}

Document recall@10 is 95.1\% on DataBench and 99.6\% on the MiMoTable retrieval track, but final accuracy is 22.3 and 17.5 points below exception-free SQL execution, respectively. Operational success is therefore a weak proxy for semantic success. In paired ablations, removing representative and sample values has the largest controlled effect, reducing gold-workbook denotation by 7.6 points and final accuracy by 7.0 points. Summaries and schema reranking contribute 2.6 and 0.9 final-accuracy points. The full cascade and ablations appear in Appendix Tables~\ref{tab:diagnostics} and~\ref{tab:ablations}.

\subsection{RQ3: Verification, repair, and controlled-input removal}

\begin{table}[H]
\centering
\small
\begin{tabular}{lrrrr}
\toprule
\textbf{Adaptation budget} & \textbf{Accepted} & \textbf{Correct} & \textbf{Overall accuracy} & \textbf{$\Delta$} \\
\midrule
Two initial candidates, no repair & 1{,}516 & 1{,}335 & 73.8\% & --- \\
$+$ first repair round & 1{,}574 & 1{,}379 & 76.2\% & $+2.4$ \\
$+$ second repair round & 1{,}576 & 1{,}381 & 76.3\% & $+2.5$ \\
\bottomrule
\end{tabular}
\caption{Paired \vigil{} repair ladder on 1{,}810 DataBench questions with a gold workbook and expected-type metadata. Abstentions count as incorrect.}
\label{tab:verifier}
\end{table}

Repair primarily increases coverage. It rises from 83.8\% with the two initial candidates to 87.1\% after two rounds. The first round adds 58 accepted outputs and 44 correct answers, or 75.9\% incremental selective accuracy. The second adds two accepted and two correct outputs. Overall accuracy rises by 2.5 points, while selective accuracy changes from 88.1\% to 87.6\%. The final cascade contains 1{,}730 executing terminal queries, 1{,}576 accepted outputs, and 1{,}381 correct answers. The 195 accepted errors show that shape and lexical checks do not establish correct filters or aggregation scope.

To remove the controlled answer-type input, we train a question-only term frequency--inverse document frequency (TF--IDF) logistic classifier. Evaluated out-of-fold with documents kept intact, it predicts the five-way contract at \OracleTypeOOFAccuracy\% accuracy. Replacing the type oracle lowers QA accuracy from \OracleGoldDocGoldTypeAccuracy\% to \OracleGoldDocPredictedTypeAccuracy\% with the gold workbook, a \OracleTypeCostGoldDoc-point cost (document-cluster bootstrap 95\% confidence interval \OracleTypeCostGoldDocLow--\OracleTypeCostGoldDocHigh). Combining predicted types with \fico{}'s frozen workbook selections yields \OracleSelectedDocPredictedTypeAccuracy\%. Using unrounded scores, this is \OracleLearnedOverZeroShotBridge{} points above a zero-shot type predictor but \OracleBridgeGap{} points below the fully controlled arm (Appendix~\ref{app:type-oracle}).

Separately, on the official MiMoTable English text-only protocol, direct GPT-5.4 obtains \MiMoDirectMean\%, while semantic-agreement gating obtains \MiMoSemanticMean\% (Table~\ref{tab:mimo-official}). The 0.05-point difference, computed from unrounded means, is descriptive and not statistically resolved. The gate promotes 28 of 45 proposed replacements. Across the 28 replacements and three judge repetitions (84 judgments), judged accuracy increases from \MiMoSemanticChangedDirect\% to \MiMoSemanticChangedAccuracy\%, a 5.00-point changed-subset gain. Because the variant was selected on the released set, the result requires hidden-test confirmation. The full variant sweep appears in Appendix~\ref{app:mimo_sweep}.

\begin{table}[H]
\centering
\small
\begin{tabular}{lr}
\toprule
\textbf{Official-workbook system} & \textbf{Mean judged accuracy} \\
\midrule
Claude~3.5 Sonnet, reported English text-only & 79.0\% \\
Direct GPT-5.4 & \MiMoDirectMean\% \\
\vigil{} semantic agreement & \MiMoSemanticMean\% \\
\bottomrule
\end{tabular}
\caption{Official MiMoTable English text-only results ($n=641$). The benchmark supplies the workbook, and GPT-4o judgments are averaged over three complete repetitions. The 0.05-point difference quoted in the text is computed from unrounded means.}
\label{tab:mimo-official}
\end{table}

\subsection{RQ4: Continual transfer and regression control}

\paragraph{Full 79-document replay.}
\begin{table}[H]
\centering
\footnotesize
\setlength{\tabcolsep}{3pt}
\begin{tabular}{lrrrrrr}
\toprule
\textbf{Treatment} & \textbf{\shortstack{Forward\\accuracy}} & \textbf{$\Delta$} & \textbf{Coverage} & \textbf{\shortstack{Selective\\accuracy}} & \textbf{\shortstack{Retrospective\\gain}} & \textbf{\shortstack{Harmful\\deployed}} \\
\midrule
Static harness & \CLStaticForward\% & --- & \CLStaticCoverage\% & \CLStaticPrecision\% & --- & --- \\
Ungated update & \textbf{\CLUngatedForward\%} & \textbf{+\CLUngatedDelta} & \textbf{\CLUngatedCoverage\%} & \textbf{\CLUngatedPrecision\%} & $+\CLUngatedRetrospective$ & $8/40$ \\
Dual replay gate & \CLGuardedForward\% & $+\CLGuardedDelta$ & \CLGuardedCoverage\% & \textbf{\CLGuardedPrecision\%} & \textbf{+\CLGuardedRetrospective} & \textbf{0/16} \\
\bottomrule
\end{tabular}
\caption{Full replay over ten five-episode orderings. Forward metrics use pre-update performance on episodes 2--5. Harmful/deployed counts include only proposals with a successor episode. Bold marks column bests, and deltas use unrounded values.}
\label{tab:continual}
\end{table}

Both adaptive treatments improve forward accuracy. Ungated updating reaches \CLUngatedForward\%, the highest forward result, while the dual replay gate reaches \CLGuardedForward\%, 0.4 points lower. The dual gate's gain over static is \CLGuardedDelta{} points with a 95\% $t$-interval of \CLGuardedDeltaLow--\CLGuardedDeltaHigh, and all ten orderings improve. Although its forward mean is lower, the dual gate has the larger retrospective gain: \CLGuardedRetrospective{} points versus \CLUngatedRetrospective{} for ungated updating. Under the dual gate, coverage rises by 6.7 points and selective accuracy rises from 87.0\% to 87.5\%. An oracle that always chooses the best answer already present in the frozen candidate pool reaches \CLActionOracle\%.

\begin{figure}[H]
\centering
\includegraphics[width=0.92\textwidth]{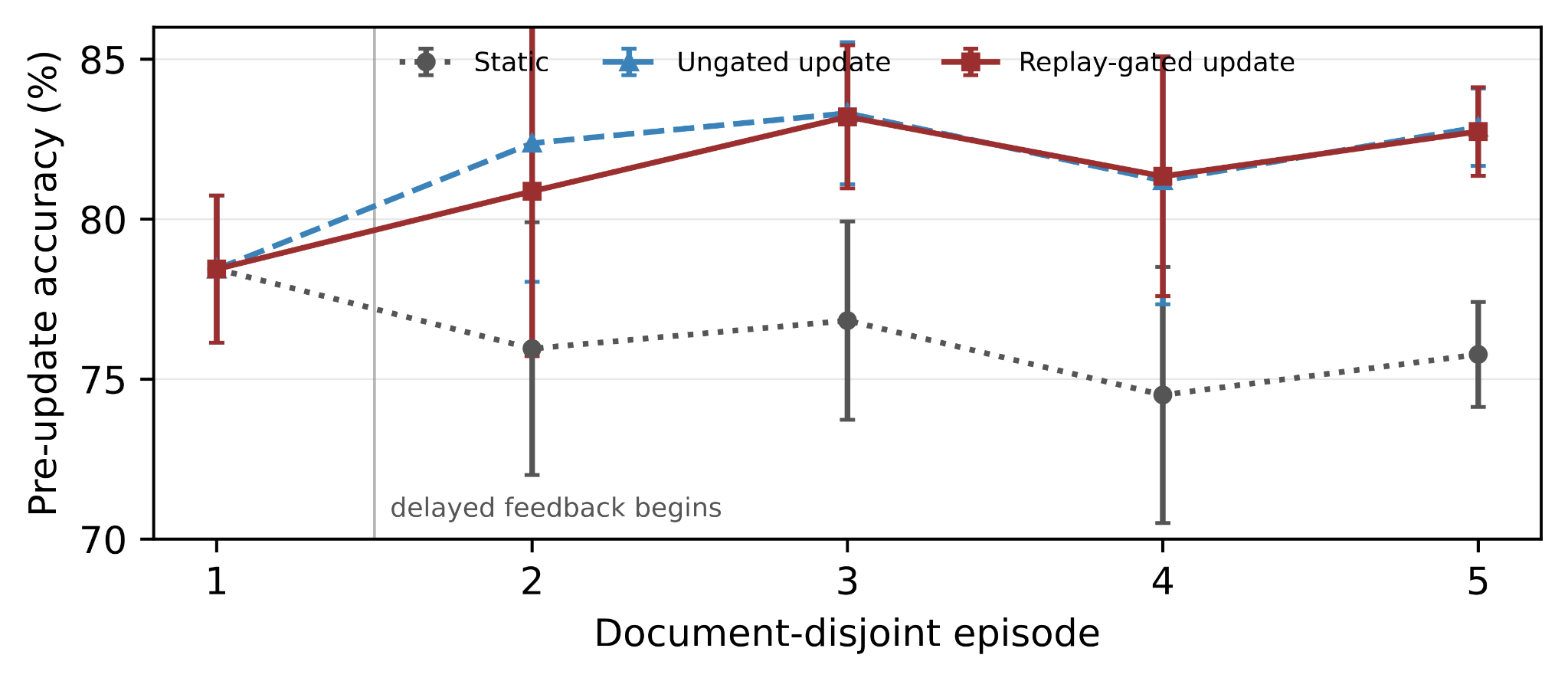}
\caption{Forward transfer during full replay. The vertical line marks the point after which feedback from episode 1 is available. Pre-update accuracy is averaged over ten document orderings, and bars show 95\% $t$-intervals. Each episode contains a different document set, so variation in the static line reflects episode difficulty rather than model change.}
\label{fig:continual}
\end{figure}

The final column of Table~\ref{tab:continual} summarizes the immediate next-episode audit. Ungated learning deploys all \CLCheckedProposals{} auditable updates, of which \CLHarmfulProposals{} are harmful. The dual gate deploys \CLCheckedPromotions{} of the same proposals, and none harms the immediate successor. Because the gate deploys fewer updates, this result demonstrates filtering rather than a causal reduction rate or safety guarantee. The local audit is distinct from the fixed held-out analysis below.

\paragraph{Fixed held-out audit.}
The stricter analysis reserves the same 16 documents and 380 questions for every ordering. They are evaluated only after each ordering's final harness is fixed.

\begin{table}[H]
\centering
\small
\setlength{\tabcolsep}{4pt}
\begin{tabular}{lrrrr}
\toprule
\textbf{Final harness} & \textbf{Accuracy} & \textbf{$\Delta$} & \textbf{Coverage} & \textbf{Selective accuracy} \\
\midrule
Static & \CLAuditStaticAccuracy\% & --- & 90.5\% & 88.7\% \\
Calibrator only & \CLAuditCalibratorAccuracy\% & $+5.9$ & 96.7\% & 89.1\% \\
Selector only & \CLAuditSelectorAccuracy\% & $+0.2$ & 90.5\% & 88.9\% \\
Combined, ungated & \CLAuditUngatedAccuracy\% & $+6.3$ & 96.7\% & 89.5\% \\
Combined, accuracy-only gate & \textbf{\CLAuditAccuracyGateAccuracy\%} & \textbf{+\CLAuditAccuracyGateDelta} & \textbf{\CLAuditAccuracyGateCoverage\%} & \textbf{\CLAuditAccuracyGatePrecision\%} \\
Combined, dual gate & \CLAuditDualGateAccuracy\% & $+6.3$ & 96.6\% & \textbf{89.6\%} \\
\bottomrule
\end{tabular}
\caption{Final performance on the fixed \CLAuditQuestions-question, \CLAuditDocuments-document audit excluded from fitting and online promotion. Values average ten ordering-specific final harnesses. Bold marks metric bests; selective accuracy is tied and deltas use unrounded values.}
\label{tab:heldout-audit}
\end{table}

The held-out audit provides evidence that learned corrections transfer beyond the reused promotion buffer. The accuracy-only gate improves accuracy by \CLAuditAccuracyGateDelta{} points, from \CLAuditStaticAccuracy\% to \CLAuditAccuracyGateAccuracy\%, while selective accuracy also rises. Calibrator-only adaptation reaches \CLAuditCalibratorAccuracy\%, a 5.9-point gain that is close to the combined 6.4-point gain. Selector-only reaches \CLAuditSelectorAccuracy\%. Gating changes the final mean little relative to combined ungated learning and remains an imperfect risk filter: \CLAuditPromotedHarmful{} of \CLAuditPromoted{} approved updates reduce held-out accuracy, while \CLAuditRejectedHarmful{} of \CLAuditRejected{} rejected updates are harmful. Gate-local non-regression therefore does not establish global or future safety~\citep{dwork2015holdout}. Appendix Table~\ref{tab:signals} summarizes recurrent verifier signals.\looseness=-1

\section{Discussion}

\paragraph{What the learner changes.}
The learner estimates which executing queries and result columns are reliable by answer type and verifier issue. The 6.4-point held-out gain provides evidence that recurring over-rejection patterns transfer across document-disjoint schemas. Calibrator-only adaptation reaches 86.2\% on the fixed audit, whereas selector-only adaptation reaches 80.5\%, indicating that most of the gain is associated with contract calibration. The frozen candidate pool enables paired attribution but cannot create a missing action when every candidate uses the wrong predicate, grouping, or workbook. Addressing semantic false acceptance therefore requires structure-aware checks, new repair prompts, regenerated candidates, or verifiers that represent predicate scope, grouping keys, set cardinality, and aggregation intent~\citep{ni2023lever}. Each change would alter the proposal distribution and require a new promotion and audit cycle.

\paragraph{Stability, evidence, and operations.}
Ungated updating has the highest forward accuracy (82.4\%), while the dual gate is 0.4 points lower. The dual gate has the larger retrospective gain (2.7 versus 2.3 points) and deploys 16 proposals with no immediate-successor harm. Ungated learning deploys all 40 proposals, including eight harmful ones. Because the treatments deploy different updates and reuse the buffer adaptively, this is not a causal safety estimate. Promotion instead acts as a risk-budget choice that trades some short-term plasticity for fewer state changes and stronger paired backtests. The primary metric and promotion constraints should be fixed prospectively.

Evidence should be interpreted in layers. Hard checks are deterministic within the implementation. The document-disjoint promotion buffer provides paired incumbent-versus-proposal evidence on observed data, while the fixed held-out audit estimates transfer beyond fitting and promotion documents on one predetermined split. Neither empirical layer establishes non-regression on future inputs or population-level safety.

The per-update audit still finds that 3 of 26 approved and 6 of 24 rejected proposals reduce held-out accuracy. Promotion reports should therefore include both approved and rejected counterfactuals, deployment rate, subgroup losses, and final-harness performance rather than only the promotion-buffer score.

Deployment records should retain trajectories, component versions, fit/gate membership, incumbent counterfactuals, and retain decisions. A promotion record can then expose changed thresholds, ranking flips, influential buffer cases, and subgroup losses. High-impact changes may require human approval or a canary phase. Equation~\ref{eq:transaction} rolls back $\theta$ exactly, but it cannot reverse external side effects. Operational reversibility also requires idempotent tools, staged writes, or compensating actions.

A longitudinal study should preserve timestamps, reveal labels only on arrival, and shadow-test proposed states. It should report harmful-update latency, rollback recovery, abstention and coverage drift, calibration, update frequency, cost, and constraint violations. Frozen replay cannot capture distribution changes after prompt or adapter updates. Any refreshed audit or change to the retriever, generator, schema adapter, or foundation model must establish a new promotion baseline and pass an independent evaluation tied to the altered action distribution.

\paragraph{Limitations.}
The episodes use fixed benchmarks, a frozen candidate set, and labels that proxy for delayed feedback. Orderings vary sequence rather than membership. The controlled verifier receives a gold workbook and expected type, while its document-disjoint type-prediction bridge is \OracleBridgeGap{} points below the fully controlled arm. The audit uses one deterministic split, and no prior method shares this spreadsheet continual protocol, so gains are within-protocol rather than state-of-the-art. On MiMoTable, Claude~3.5 Sonnet is reported rather than rerun, GPT-5.4 is newer, GPT-4o judging is not human adjudication, the 0.05-point difference is unresolved, and variant selection is test-adaptive. External-task results cover only three domains and report only five-run mean reward. We do not report per-question latency distributions, token cost, or indexing cost. Cross-sheet joins, formulas, charts, and editing remain unsupported.

\section{Conclusion}

\vigil{} makes the harness, not the foundation model, the continual learner. Delayed feedback raises forward accuracy from 75.8\% to 82.0\% in full replay and held-out accuracy from \CLAuditStaticAccuracy\% to \CLAuditAccuracyGateAccuracy\%, with most held-out gain attributable to contract calibration. Three of 26 gate-approved updates nevertheless reduce held-out accuracy, showing that finite replay gates are empirical controls rather than safety guarantees. The practical design lesson is to keep mutable state narrow, preserve replayable incumbent counterfactuals, promote against document-disjoint evidence, and audit final states on independent data. Stronger semantic verification, calibrated abstention, regenerated-candidate studies, and end-to-end evaluation without workbook or type oracles remain necessary before deployment.

\clearpage
\bibliographystyle{plainnat}
\bibliography{vigil_references}

\clearpage

\appendix

\section{Web-Page Control Experiment}
\label{app:html}

We use a corpus of web pages as a control setting in which answers are more often stated in text. The corpus contains 44 pages from four domains and 190 question--answer pairs: detail lookup (91), cross-section (36), multi-hop (18), presence/absence (25), and summary (20). Raw page text is extracted with Trafilatura, split into 350-token non-overlapping chunks, embedded with \texttt{embeddinggemma-300m}, and retrieved at depth ten.

\begin{table}[H]
\centering
\small
\begin{tabular}{lcc}
\toprule
\textbf{Outcome} & \textbf{Count} & \textbf{Share} \\
\midrule
Correct & 103 & 54\% \\
Retrieval within document & 59 & 31\% \\
Retrieval across documents & 26 & 14\% \\
Extraction error & 2 & 1\% \\
Generation error & 0 & 0\% \\
\bottomrule
\end{tabular}
\caption{Web-page control results. In this audited corpus, retrieval errors dominate after extraction.}
\label{tab:html}
\end{table}

This control illustrates that fixed-size text chunks can be adequate when the answer is explicitly stated. It does not by itself establish that 350-token chunks are optimal for web pages or that the same error profile holds in other corpora.

\section{Prefix-RAG Baseline Details}
\label{app:baseline}

The baseline parser detects sheet boundaries, scores candidate header rows, and retains up to three sample rows per sheet. The resulting text is truncated to 2{,}000 tokens, embedded as one document-level unit, and retrieved as context for the generator. Because the full table is not available at answer time, global aggregation usually cannot be computed.

For example, for the 2{,}668-row Forbes billionaires dataset, the question ``How many billionaires are under 40?'' is answered from three sample rows by the baseline. \fico{} instead executes:

\begin{nolinenumbers}
\begin{lstlisting}[language=SQL]
SELECT COUNT(*) FROM data
WHERE age < 40 AND age IS NOT NULL;
-- Result: 87
\end{lstlisting}
\end{nolinenumbers}

\section{Diagnostic Protocol and Coverage Argument}
\label{app:coverage}

Coverage@5 is annotated as positive only when the five retrieved units jointly expose the values or sufficient statistics needed for the reference computation. For an aggregation whose result can change when an unseen row changes, context that omits that row and provides no sufficient statistic cannot uniquely determine the answer. Precomputed summaries can increase coverage for anticipated operations, but they do not guarantee answers to arbitrary filters or groupings.

The diagnostic is specific to its corpus, representation choices, retrieval depth, and annotation protocol. Curves over $k$, context length, row count, and question type would be needed to estimate how coverage changes with budget.

\begin{table}[H]
\centering
\small
\begin{tabular}{lcc}
\toprule
\textbf{Representation} & \textbf{Coverage@5} & \textbf{End-to-End} \\
\midrule
350-token chunks & 16\% & 5\% \\
Row packing & 29\% & 10\% \\
Precomputed summaries & 39\% & 15\% \\
Multi-granularity~\citep{gulati2026frtr} & 6\% & 6\% \\
\midrule
Whole-table execution & --- & \textbf{78\%} \\
\bottomrule
\end{tabular}
\caption{Diagnostic results under a fixed top-5 retrieval budget. Coverage@5 is whether retrieved text contains the evidence needed to answer. Whole-table execution bypasses answer-from-chunks.}
\label{tab:chunking}
\end{table}

\section{Method Details}
\label{app:method}

\paragraph{Document summaries.}
The summary prompt asks for the document topic, the contents of each sheet, verbatim key column names, and example questions the data can answer. A representative summary is:

\begin{quote}
\small\itshape
``This dataset tracks 2{,}668 billionaires from the Forbes World's Billionaires list. It covers net worth, age, self-made status, gender, philanthropy score, and country of origin. Key columns: personName, finalWorth, age, selfMade, gender, philanthropyScore, country. It can answer questions about wealth, country distributions, age breakdowns, and wealth-source categories.''
\end{quote}

\paragraph{Schema prompt.}
A representative execution prompt includes:

\begin{nolinenumbers}
\begin{lstlisting}
Table: data (2,668 rows)
Columns:
  personName: text, e.g. Elon Musk, Jeff Bezos
  finalWorth: numeric, range 1,000 to 219,000
  age: numeric, range 19 to 96
  selfMade: boolean, e.g. True, False
  gender: text, e.g. M, F
  country: text, e.g. United States, China
Sample rows:
  Elon Musk | 219000 | 50 | True | M | United States
  Jeff Bezos | 171000 | 58 | True | M | United States
\end{lstlisting}
\end{nolinenumbers}

Generation rules cover null handling, date comparisons, group comparisons, threshold counts, and derived rates. The result is passed to the answer model with instructions to treat the database output as authoritative and preserve exact values.

\section{Oracle, Verifier, and Ablation Reporting Details}
\label{app:oracle}

The gold-workbook run changes only the workbook supplied to the compute stage. It does not expose the gold answer, and it retains the same loader, schema prompt, generation rules, retry policy, answer synthesis, and judge. The benchmark does not provide gold sheet labels, so the oracle retains sheet selection within each workbook.

Denotation correctness is evaluated on a canonical result object before free-form synthesis. We retain all questions in the denominator, including parse failures, contract rejections, and execution errors. Ablation effects use paired question sets, and execution and semantic correctness are reported separately.

Empty-result fallback triggers on 14 questions (0.8\%), repairing one final answer and harming one. The remaining 12 triggered cases are unchanged. Removing representative and sample values reduces gold-workbook denotation accuracy from \FicoDBGoldDenotation{} to \FicoDBGoldNoSamplesDenotation{} and final accuracy from \FicoDBGoldFinal{} to \FicoDBGoldNoSamplesFinal{}.

\paragraph{Verifier-run reproducibility.}
The full verifier treatment used 12 workers, two diversely prompted initial candidates, two possible repair rounds, and a fresh output directory with per-question checkpoints. The 1{,}810-question workflow completed successfully in 38 minutes 23 seconds. Persisted artifacts include the selected query and result, all candidate queries and denotations, verification issues, selected stage, typed target values, and deterministic correctness. No benchmark gold answer is included in generation, verification, repair feedback, or selection. It is read only by the evaluator after selection. The controlled verifier and replay tracks do receive the five-way expected answer type, which is therefore listed as an oracle in Table~\ref{tab:ledger}.

\subsection{Expected-Type Oracle Removal}
\label{app:type-oracle}

We replace the expected-type field with a question-text-only classifier using word 1--3-gram and character 3--5-gram term frequency--inverse document frequency (TF--IDF) features and class-balanced logistic regression. Ten-fold GroupKFold keeps all questions from a document together: every predicted contract is produced by a model trained without that question's document. The resulting out-of-fold type accuracy is \OracleTypeOOFAccuracy\% over all 1{,}810 questions. A separate model fitted on the 63 continual-learning documents reaches \OracleTypeAuditAccuracy\% on the fixed 16-document held-out audit. These are supervised benchmark-transfer estimates, not zero-label deployment results.

The $2\times2$ experiment crosses gold versus frozen \fico{}-selected workbooks with gold versus out-of-fold predicted types (Table~\ref{tab:oracle-removal}). Candidate generation and verification receive the assigned workbook and contract type but never the gold answer. The gold answer is read only after selection for deterministic scoring. In predicted-type arms, the gold type is also used only for evaluation. In gold-type arms, it is the controlled input specified in Table~\ref{tab:ledger}. Using unrounded scores, the learned classifier improves the fully bridged arm by \OracleLearnedOverZeroShotBridge{} points over a question-only zero-shot classifier, whose type accuracy is \OracleZeroShotTypeAccuracy\% and bridged QA accuracy is \OracleZeroShotBridgeAccuracy\%. Nevertheless, incorrectly predicted types remain consequential: only 15 of 183 such questions are answered correctly with the gold workbook.

\begin{table}[H]
\centering
\small
\begin{tabular}{lrr}
\toprule
\textbf{Workbook supplied} & \textbf{Gold type} & \textbf{Predicted type} \\
\midrule
Gold workbook & \OracleGoldDocGoldTypeAccuracy\% & \OracleGoldDocPredictedTypeAccuracy\% \\
\fico{}-selected workbook & \OracleSelectedDocGoldTypeAccuracy\% & \OracleSelectedDocPredictedTypeAccuracy\% \\
\bottomrule
\end{tabular}
\caption{Deterministic QA accuracy in the workbook/type oracle-removal experiment. The fully controlled-to-bridge gap is \OracleBridgeGap{} points (document-cluster bootstrap 95\% confidence interval \OracleBridgeGapLow--\OracleBridgeGapHigh).}
\label{tab:oracle-removal}
\end{table}

\paragraph{Post-hoc rejection audit.}
For each abstention, we inspect the selected rejected query's raw denotation and test each result column against the gold answer using the same typed equality rules. This finds 121 answer-containing denotations among 234 abstentions. Of these, 38 are category or list-of-category compatibility failures and 69 trigger the rule requiring explicit \texttt{COUNT} for count questions. The procedure is intentionally diagnostic: choosing a matching column with the gold answer is an oracle operation, and denotation equality may be coincidental, so these cases cannot be counted as deployable accuracy. They nevertheless show that both column compatibility and lexical intent rules require calibration.

\paragraph{Trace-signal aggregation.}
The 4{,}150 candidate records retain separate \texttt{error} and \texttt{issues} fields. Table~\ref{tab:signals} groups exact parser and binder error prefixes and normalized verifier messages. Counts are candidate-event counts, not unique questions. One candidate can emit multiple issues. Empty messages and diagnostic line excerpts are excluded.

\begin{table}[H]
\centering
\small
\begin{tabular}{lrl}
\toprule
\textbf{Observed signal} & \textbf{Events} & \textbf{Potential remediation} \\
\midrule
Count intent but no \texttt{COUNT} & 293 & Intent representation / generation prompt \\
Parser error & 236 & Structured decoding / repair examples \\
Type or result-shape mismatch & 193 & Contract calibration / result-column selection \\
Binder error & 169 & Schema normalization / grounding \\
Read-only or single-query violation & 122 & Constrained decoding \\
\bottomrule
\end{tabular}
\caption{Frequent verifier signals across candidate trajectories. Signals are non-exclusive event counts, not question-level error rates. Only contract calibration and query/result-column selection are learned in the continual experiment.}
\label{tab:signals}
\end{table}

\paragraph{Continual-replay reproducibility.}
The replay experiment uses five episodes and seeds 0--9. The trace store contains 79 document identifiers covering all 1{,}810 question trajectories. Documents are shuffled per seed and greedily assigned to the currently smallest episode by question count. No document identifier appears in two episodes. Within observed episodes, a stable document hash assigns 70\% of documents to fitting and 30\% to the promotion gate. The selector is a regularized logistic model over answer type, initial/repair stage, query-length bin, SQL operator indicators and their answer-type interactions, verifier issue class, result-column position, and lexical column cues. A fitting-only sweep chooses the relaxed-contract threshold subject to no more than a 0.5-point selective-accuracy loss. Promotion then requires positive accuracy gain and no selective-accuracy loss on the disjoint gate buffer.

For the held-out-audit experiment, the 16 lowest document hashes under a separate fixed audit salt define an immutable 380-question held-out audit. The other 63 documents and 1{,}430 questions form the sequential learning set. The audit is evaluated only after each ordering's final harness is fixed. Six treatments comprise the static harness, calibrator-only, selector-only, their ungated combination, accuracy-only promotion, and accuracy-plus-selective-accuracy promotion. The static audit score is sequence invariant. Adaptive means reflect ten ordering-specific final harnesses evaluated on the same held-out questions.

For rejected candidates that execute and remain read-only, the calibrator may consider type-compatible result columns. It cannot relax parser, binder, execution, 500-row-cap, multiple-statement, mutation, or read-only failures. Gold answers label actions only after their episode and never enter action features. A next-episode proposal is classified as harmful only when it lowers deterministic accuracy relative to the pre-update harness. This audit is available for episodes 1--4 in each ordering, not for final-episode proposals. The static replay exactly reproduces 1{,}576 accepted and 1{,}381 correct outputs from the verifier run. The replay is fully determined by the persisted candidate records, document-order seed, fit/gate hash split, feature specification, and promotion rule.

\section{Extended Results}
\label{app:extended}

\begin{table}[htbp]
\centering
\small
\begin{tabular}{lcc}
\toprule
\textbf{Diagnostic} & \textbf{DataBench} & \textbf{MiMoTable} \\
\midrule
Document recall@10 & 95.1\% & 99.6\% \\
Gold document selected for execution & 81.3\% & 90.9\% \\
SQL execution without exception & 98.5\% & 97.2\% \\
Final answer correct & 76.2\% & 79.7\% \\
\bottomrule
\end{tabular}
\caption{Deployed \fico{} stage cascade. High retrieval and execution rates do not imply final-answer correctness. All rows belong to the end-to-end track.}
\label{tab:diagnostics}
\end{table}

\begin{table}[htbp]
\centering
\small
\setlength{\tabcolsep}{4pt}
\begin{tabular}{llrrr}
\toprule
\textbf{Stage} & \textbf{Intervention} & \textbf{Full} & \textbf{Ablated} & \textbf{$\Delta$} \\
\midrule
Find & Remove summaries (final) & \FicoDBFullFinal & \FicoDBNoSummaryFinal & $-2.6$ \\
Find & Remove schema reranking (final) & \FicoDBFullFinal & \FicoDBNoRerankFinal & $-0.9$ \\
Find & Remove empty-result fallback (final) & \FicoDBFullFinal & \FicoDBNoFallbackFinal & $0.0$ \\
Compute & Remove representative/sample values (denotation) & \FicoDBGoldDenotation & \FicoDBGoldNoSamplesDenotation & $-7.6$ \\
Compute & Remove representative/sample values (final) & \FicoDBGoldFinal & \FicoDBGoldNoSamplesFinal & $-7.0$ \\
\bottomrule
\end{tabular}
\caption{Paired DataBench ablations in a fixed controlled runner. Compute rows use the gold workbook. Only within-row deltas are interpreted.}
\label{tab:ablations}
\end{table}

\begin{table}[H]
\centering
\small
\setlength{\tabcolsep}{5pt}
\begin{tabular}{lrrr}
\toprule
\textbf{Answer type} & \textbf{Questions} & \textbf{Correct} & \textbf{Accuracy} \\
\midrule
Boolean & 387 & 328 & 84.8\% \\
Number & 412 & 322 & 78.2\% \\
Category & 333 & 240 & 72.1\% \\
List[category] & 329 & 210 & 63.8\% \\
List[number] & 349 & 281 & 80.5\% \\
\bottomrule
\end{tabular}
\caption{Gold-workbook \vigil{} accuracy by expected answer type. The expected type is controlled metadata in this track.}
\label{tab:bytype}
\end{table}

\begin{table}[H]
\centering
\small
\begin{tabular}{lrrr}
\toprule
\textbf{Terminal candidate stage} & \textbf{Questions} & \textbf{Correct} & \textbf{Accuracy} \\
\midrule
Initial candidate 1 & 1{,}456 & 1{,}157 & 79.5\% \\
Initial candidate 2 & 258 & 178 & 69.0\% \\
Repair 1 & 89 & 44 & 49.4\% \\
Repair 2 & 7 & 2 & 28.6\% \\
\bottomrule
\end{tabular}
\caption{Verifier treatment by terminal candidate stage, including accepted and rejected terminal candidates. Repair rows are conditioned on both initial candidates failing and are therefore harder. Accuracies are descriptive, not causal effects.}
\label{tab:verifierstage}
\end{table}

\begin{table}[H]
\centering
\small
\begin{tabularx}{\textwidth}{Y Y Y}
\toprule
\textbf{Question} & \textbf{Baseline} & \textbf{\fico{}} \\
\midrule
``Among the top five countries by reported coronavirus cases, which has the highest case-fatality ratio?'' & ``I don't know. The context only lists the United States.'' & Brazil (2.189\%) from a grouped ratio query. \\
\midrule
``Among non-drinkers, is gambling mostly Yes or No?'' & ``Based on context, I cannot determine.'' & Mostly No: 89 No versus 33 Yes. \\
\midrule
``Do PASS or FAIL movies have a higher profitable share?'' & ``I don't know. The context does not contain financial data.'' & PASS: 64.2\% versus FAIL: 61.8\%. \\
\bottomrule
\end{tabularx}
\caption{Representative cases in which full-table execution answers an aggregation that the prefix context cannot support.}
\label{tab:hero}
\end{table}

\section{Official MiMoTable Variant Sweep}
\label{app:mimo_sweep}

\begin{table}[H]
\centering
\small
\setlength{\tabcolsep}{4pt}
\begin{tabularx}{\textwidth}{lYrr}
\toprule
\textbf{System} & \textbf{Selection rule} & \textbf{Mean (\%)} & \textbf{$\Delta$ vs. direct (points)} \\
\midrule
Direct GPT-5.4 & Official prompt, medium reasoning & 85.66 & --- \\
One verifier & Workbook-aware replace/retain & 85.58 & $-0.08$ \\
Two-verifier consensus & Second verifier authorizes replacement & 85.59 & $-0.07$ \\
Three-candidate deliberation & Randomized direct/calculation/audit arbitration & 85.51 & $-0.15$ \\
Calculation candidate & Calculation-specialized prompt & 84.96 & $-0.70$ \\
Audit candidate & Audit-specialized prompt & 84.70 & $-0.96$ \\
Candidate agreement & Calculation and audit answers agree & 85.32 & $-0.34$ \\
Exact prompt, high reasoning & No gate, reasoning-effort control & 85.59 & $-0.07$ \\
Semantic agreement & Verifier correction agrees with exact-prompt solve & 85.70 & $+0.05$ \\
Three-signal intersection & Semantic agreement plus second authorization & 85.46 & $-0.20$ \\
\bottomrule
\end{tabularx}
\caption{Complete gold-blind MiMoTable sweep. Every mean averages three complete judgments by the official GPT-4o prompt. The 0.05-point semantic-agreement difference is descriptive and not statistically resolved. Choosing this variant after the released-test sweep is test-adaptive.}
\label{tab:mimo_sweep}
\end{table}

Semantic agreement promotes 28 of 45 verifier proposals. On those 28 questions, mean judged accuracy rises from 61.43\% to 66.43\%. This changed-subset result is diagnostic rather than a full-system estimate. A second diagnostic reuses direct judgments for the 613 unchanged predictions and gives 85.88\%, but it is not the official full-system protocol. The three-signal intersection promotes 11 answers and improves that changed subset by 18.18 points. Its separately rerun full-system mean is lower because the stochastic judge shifts on byte-identical answers. These diagnostics motivate paired or cached judging in future benchmark designs.

\section{Cross-Domain Promote-or-Retain Case Studies}
\label{app:clbench}

We also instantiate the same evidence-gated state-update discipline in three domains from Continual Learning Bench (CL-Bench)~\citep{asawa2026clbench}, using commit \texttt{5f8c50e}, each task's default schedule, five permuted runs, and the official scorer. These are task-specific adapters rather than one shared learned state. The spectrum adapter promotes persistent transmitters only after repeated observations satisfy support and geometric constraints. The sales adapter retains bounded history and promotes a trend model only when its rolling backtest is non-regressive. The poker adapter uses disclosed opponent identities and deterministic Monte Carlo equity. The adapters make no model calls.

\begin{table}[H]
\centering
\small
\begin{tabular}{lrrr}
\toprule
\textbf{CL-Bench domain} & \textbf{Highest released mean} & \textbf{Adapter mean} & \textbf{$\Delta$} \\
\midrule
Blind spectrum monitoring & \CLBSpectrumPrior & \CLBSpectrumVigil & $+\CLBSpectrumDelta$ \\
Sales prediction & \CLBSalesPrior & \CLBSalesVigil & $+\CLBSalesDelta$ \\
Exploitable poker & \CLBPokerPrior & \CLBPokerVigil & $+\CLBPokerDelta$ \\
\bottomrule
\end{tabular}
\caption{Mean total reward over five official-schedule runs per domain. The comparator is the highest five-run mean in the released artifacts at the pinned commit. These descriptive comparisons do not extend to other metrics.}
\label{tab:clbench}
\end{table}

The adapters attain higher observed five-run mean total reward than the released artifacts at the pinned commit by 9.407, 1.129, and 3.380 reward units, respectively. Spectrum and sales illustrate evidence- or backtest-gated state changes. Poker raises mean total reward but has zero cumulative gain, so it demonstrates a bounded task policy rather than positive continual learning. We do not report uncertainty beyond the five-run means, and these case studies do not establish a general cross-domain learner.

\section{Additional Limitations and Deployment Considerations}
\label{app:limitations}

\paragraph{Production equivalence.}
The local parser approximates the deployed system. Implementation differences can change accuracy.

\paragraph{Latency and cost.}
The verifier workflow completed 1{,}810 questions with 12 workers in 38 minutes 23 seconds, including orchestration. It uses multiple model calls. We did not measure per-question latency quantiles, token cost, or indexing cost.

\paragraph{Evaluation validity.}
MiMoTable references may not match terse database outputs, so human adjudication is needed before treating the score as a strict lower bound. The end-to-end and controlled tracks use different evaluators, and the controlled tracks receive expected-type metadata. Dataset-clustered intervals would also be preferable to question-level uncertainty alone.

\paragraph{Security, privacy, and execution language.}
The runners require each candidate to be one query beginning with \texttt{SELECT} or \texttt{WITH}, reject embedded semicolons and denylisted patterns, cap results at 500 rows, and bound prompts. These lexical checks are not a complete SQL parser. Execution has no explicit timeout, and cell-level prompt injection is not mitigated. Python or pandas code can express operations that are awkward in SQL, but arbitrary code execution has a larger attack surface. A controlled comparison with sandboxed dataframe operations is left for future work. The controlled environment does not establish production security or safety.

\end{document}